\documentclass[11pt,a4paper]{article}
\usepackage[hyperref]{acl2020}
\usepackage{times}
\usepackage{latexsym}

\usepackage{booktabs}
\usepackage{array}
\usepackage{graphicx}
\usepackage{amsmath}
\usepackage{bbm}
\usepackage{enumitem} 

\usepackage{color}
\usepackage{xcolor}
\definecolor{dartmouthgreen}{rgb}{0.05, 0.5, 0.06}

\newcommand{\ignore}[1]{}

\usepackage{microtype}

\aclfinalcopy 

\title{Text and Causal Inference: \\
A Review of Using Text to Remove Confounding from Causal Estimates
}

\author{Katherine A. Keith, David Jensen, {\rm and} Brendan O'Connor \\
  College of Information and Computer Sciences\\
  University of Massachusetts Amherst\\
  \texttt{\{kkeith,jensen,brenocon\}@cs.umass.edu}
  }

\date{}

\begin{document}
\maketitle
\begin{abstract}
Many applications of computational social science aim to infer causal conclusions from non-experimental data. Such \textit{observational} data often contains \textit{confounders}, variables that influence both potential causes and potential effects. Unmeasured or \textit{latent} confounders can bias causal estimates, and this has motivated interest in measuring potential confounders from observed text. For example, an individual’s entire history of social media posts or the content of a news article could provide a rich measurement of multiple confounders.
Yet, methods and applications for this problem are scattered across different communities and evaluation practices are inconsistent.
This review is the first to gather and categorize these examples and provide a guide to data-processing and evaluation decisions. Despite increased attention on adjusting for confounding using text, there are still many open problems, which we highlight in this paper. 
\end{abstract}

\section{Introduction}
In contrast to descriptive or predictive tasks, causal inference aims to understand how \emph{intervening} on one variable affects another variable \cite{holland1986statistics,pearl2000causality,morgan2015counterfactuals,imbens2015causal,hernan2020}. Specifically, many applied researchers aim to estimate the size of a specific causal effect,
the effect of a single \emph{treatment} variable on an \emph{outcome} variable. However, a major challenge in causal inference is addressing \emph{confounders}, variables that influence both treatment and outcome.
For example, consider estimating the size of the causal effect of smoking (treatment) on life expectancy (outcome). 
Occupation is a potential confounder that may influence both the propensity to smoke and life expectancy. Estimating the effect of treatment on outcome without accounting for this confounding could result in strongly biased estimates and thus invalid causal conclusions. 

\begin{figure}[t!]
\centering
\includegraphics[width=0.9\columnwidth]{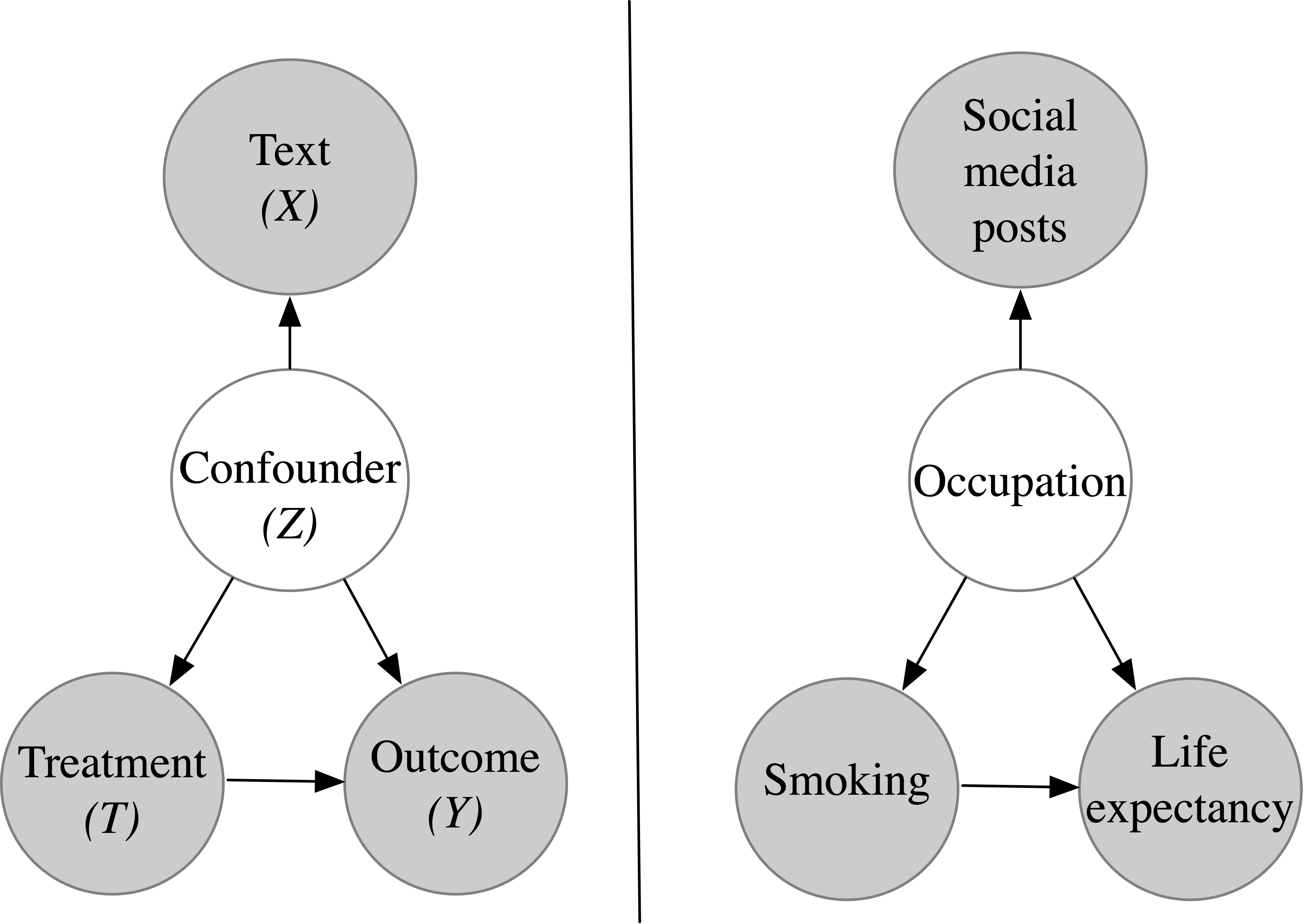}
\caption{\emph{Left:} A causal diagram for text that encodes causal confounders, the setting that is focus of this review paper.  The major assumption is that latent confounders can be \textit{measured} from text and those confounder measurements can be used in causal adjustments. 
\emph{Right:} An example application in which practitioner does not have access to the confounding variable, \emph{occupation}, in structured form but can measure confounders from unstructured text (e.g.~an individual's social media posts). 
\label{f:hook-fig}}
\end{figure}

To eliminate confounding bias, one approach is to perform randomized controlled trials (RCTs) in which researchers randomly assign treatment. Yet, in many research areas such as healthcare, education, or economics, randomly assigning treatment is either infeasible or unethical. For instance, in our running example, one cannot ethically randomly assign participants to smoke since this could expose them to major health risks. In such cases, researchers instead use observational data and adjust for the confounding bias  statistically with methods such as matching, propensity score weighting, or regression adjustment (\S\ref{s:estimators}). 

In causal research about human behavior and society, there are potentially many latent confounding variables that can be measured from unstructured text data.
Text data could either (a) serve as a surrogate for potential confounders; or (b) the language of text itself could be a confounder. 
Our running example is an instance of text as a surrogate: a researcher may not have a record of an individual's occupation but could attempt to measure this variable from the individual's entire history of social media posts (see Fig.~\ref{f:hook-fig}). An example of text as a direct confounder: the linguistic content of social media posts could influence censorship (treatment) and future posting rates (outcome) \cite{roberts2018adjusting}.

A challenging aspect of this research design is the high-dimensional nature of text. 
Other work has explored general methods for adjusting for high-dimensional confounders \cite{d2017overlap, rassen2011covariate,louizos2017causal,li2016matching,athey2017estimating}.
 However, text data differ from other high-dimensional data-types because intermediate confounding adjustments can be read and evaluated by humans (\S\ref{sec:eval-adjust}) and designing meaningful representations of text is still an open research question.\footnote{For instance, there have been four workshops on representation learning at major NLP conferences in the last four years \cite{ws-2016-representation, ws-2017-representation, ws-2018-representation, ws-2019-representation}.}
Even when applying simple adjustment methods, a practitioner must first transform text into a lower-dimensional representation via, for example, filtered word counts, lexicon indicators, topic models, or embeddings (\S\ref{s:represent}). 
An additional challenge is that empirical evaluation in causal inference is still an open research area \cite{dorie2019automated,gentzel2019case} and text adds to the difficulty of this evaluation (\S\ref{s:eval-est}).

We narrow the scope of this paper to review methods and applications with text data as a causal \emph{confounder}. In the broader area of text and causal inference, work has examined
text as a mediator \cite{veitch2019using}, text as treatment \cite{fong2016discovery,egami2018make,wood2018challenges,tan2014effect}, text as outcome \cite{egami2018make}, causal discovery from text \cite{mani2000causal}, and predictive (Granger) causality with text \cite{balashankar2019identifying,del2016case,tabari-etal-2018-causality}.

\begin{figure*}[t!]
\centering
\includegraphics[width=0.7\textwidth]{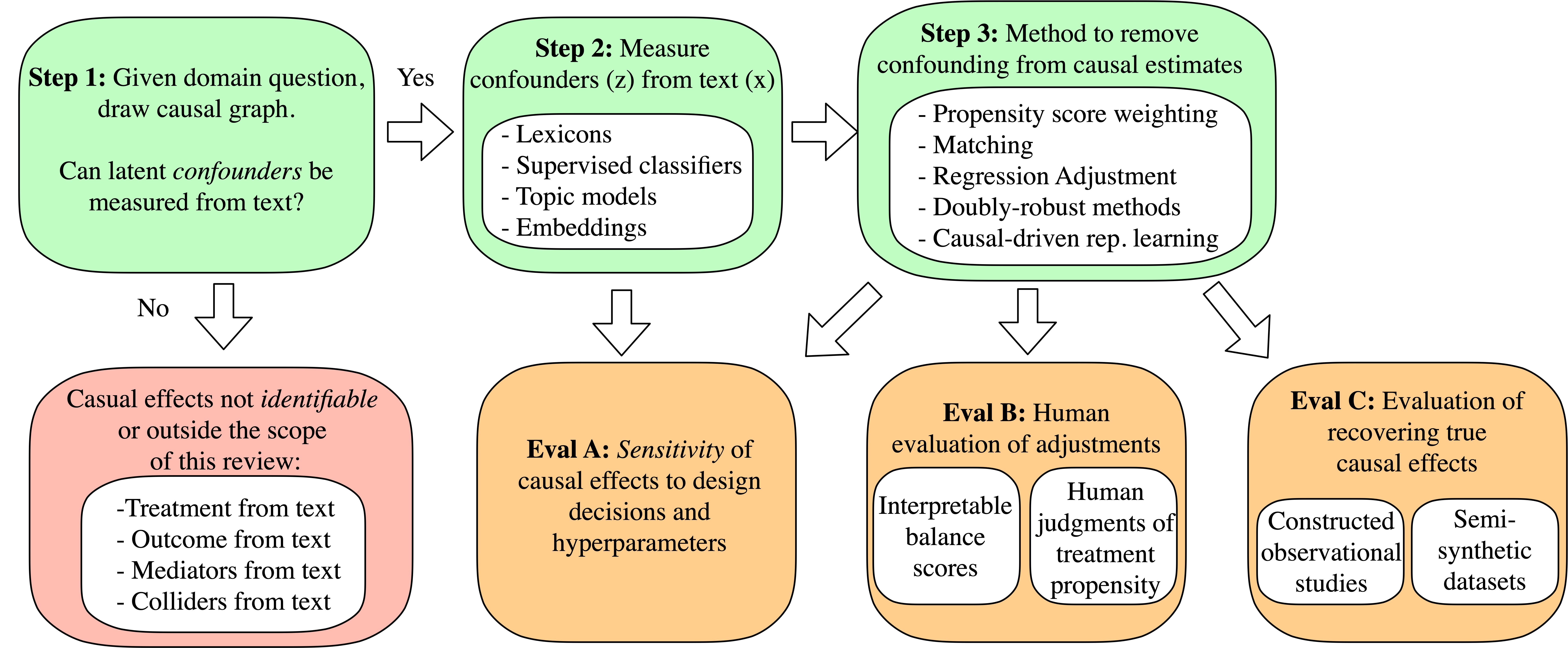}
\caption{This chart is a guide to design decisions for applied research with causal confounders from text. \emph{Step 1:} Encode domain assumptions by drawing a causal diagram (\S\ref{s:causal-intro}). If the application does not use text to measure latent \emph{confounders}, the causal effects are not identifiable or the application is outside the scope of this review. \emph{Step 2:} Use NLP to measure confounders from text (\S\ref{s:represent}). \emph{Step 3:} Choose a method that adjusts for confounding in causal estimates (\S\ref{s:estimators}). Evaluation should include \emph{(A)} sensitivity analysis (\S\ref{s:represent}), \emph{(B)} human evaluation of adjustments when appropriate (\S\ref{sec:eval-adjust}), and \emph{(C)} evaluation of recovering the true causal effects (\S\ref{s:eval-est}). 
\label{f:flow}}
\end{figure*}

Outside of this prior work, there has been relatively little interaction between natural language processing (NLP) research and causal inference. NLP has a rich history of applied modeling and diagnostic pipelines that causal inference could draw upon. Because applications and methods for text as a confounder have been scattered across many different communities, this review paper aims to gather and unify existing approaches and to concurrently serve three different types of researchers and their respective goals: 
\begin{itemize}[leftmargin=*]
    \itemsep0em
    \item \textbf{For applied practitioners,} we 
    collect and categorize applications with text as a causal confounder (Table~\ref{tab:examples} and \S\ref{s:examples}), and we provide a flow-chart of analysts' decisions for this problem setting (Fig.~\ref{f:flow}). 
     \item \textbf{For causal inference researchers working with text data,}
     we highlight recent work in representation learning in NLP (\S\ref{s:represent}) and caution that this is still an open research area with questions of the sensitivity of effects to choices in representation. We also outline existing interpretable evaluation methods for adjustments of text as a causal confounder (\S\ref{sec:eval-adjust}). 
    \item \textbf{For NLP researchers working with causal inference,} 
    we summarize some of the most-used causal estimators that condition on confounders: matching, propensity score weighting, regression adjustment, doubly-robust methods, and causally-driven representation learning (\S\ref{s:estimators}). We also discuss evaluation of methods with constructed observational studies and semi-synthetic data (\S\ref{s:eval-est}). 
\end{itemize}

\begin{table*}[t!]
    \centering
    \tiny
    \begin{tabular}{
    >{\raggedright\arraybackslash}p{1.7cm} 
    >{\raggedright\arraybackslash}p{2.0cm} 
    >{\raggedright\arraybackslash}p{2.0cm} 
    >{\raggedright\arraybackslash}p{2.2cm} 
    >{\raggedright\arraybackslash}p{1.8cm} 
    >{\raggedright\arraybackslash}p{1.5cm} 
     >{\raggedright\arraybackslash}p{1.9cm} 
    >{\raggedright\arraybackslash}p{1.9cm} 
    }
    \toprule
    
         \textbf{Paper} &  \textbf{Treatment} & \textbf{Outcome(s)} & \textbf{Confounder} & \textbf{Text data} & \textbf{Text rep.} & \textbf{Adjustment method}
          \\
         \toprule
         \citet{johansson2016learning} 
         & Viewing device (mobile or desktop) 
         & Reader's experience & News content 
         & News 
         & Word counts & Causal-driven rep. learning
         \\
         \hline 
         \citet{de2016discovering}  & Word use in mental health community & User transitions to post in suicide community & Previous text written in a forum
         & Social media (Reddit)
         & Word counts & Stratified propensity score matching
         \\
         \hline 
         \citet{de2017language} &
          
         Language of comments & 
          User transitions to post in suicide community &
          User's previous posts and comments received &
          Social media (Reddit) & Unigrams and bigrams & 
         Stratified propensity score matching
         \\
         \hline 
         \citet{falavarjani2017estimating} &  Exercise (Foursquare checkins) & Shift in topical interest on Twitter & Pre-treatment topical interest shift & 
         Social media (Twitter, Foursquare) &
         Topic models & Matching
         \\
          \hline 
         \citet{olteanu2017distilling} & Current word use & Future word use & Past word use & 
         Social media (Twitter) &
         Top unigrams and bigrams & Stratified propensity score matching 
         \\
         \hline 
         \citet{pham2017deep} & Group vs.~individual loan requests & Time until borrowers get funded & Loan description & 
         Microloans (Kiva) & 
         Pre-trained embeddings + neural networks & A-IPTW, TMLE
         \\
         \hline 
         \citet{kiciman2018using} &  Alcohol mentions & College success (e.g.~study habits, risky behaviors, emotions) & 
         Previous posts & 
         Social media (Twitter) &
         Word counts & Stratified propensity score matching
         \\
          \hline 
         \citet{sridhar2018mood} &  Exercise & Mood &  Mood triggers & 
         Users' text on mood logging apps &
         Word counts & Propensity score matching\\
         \hline
         \citet{saha2019social} & Self-reported usage of psychiatric medication & Mood, cognition, depression, anxiety, psychosis, and suicidal ideation & Users' previous posts &  Social media (Twitter) & Word counts + lexicons + supervised classifiers & Stratified propensity score matching
         \\
         \hline
         \citet{sridhar2019estimating} & Tone of replies & Changes in sentiment & Speaker's political ideology & Debate transcripts & Topic models + lexicons  & Regression adjustment, IPTW, A-IPTW\\
         \hline 
         \citet{veitch2019using} &  Presence of a theorem & Rate of acceptance & Subject of the article& Scientific articles & BERT & Causal-driven rep. learning + Regression adjustment, TMLE
         \\
         \hline
         \citet{roberts2018adjusting} &  Perceived gender of author & Number of citations & Content of article & 
         International Relations articles &
         Topic models + propensity score & Coarsened exact matching 
         \\
         \hline
         \citet{roberts2018adjusting} &  Censorship & Subsequent censorship and posting rate & Content of posts & Social media (Weibo) & Topic models + propensity score & Coarsened exact matching \\
         \bottomrule
    \end{tabular}
    \caption{Example applications that infer the causal effects of treatment on outcome by measuring confounders (unobserved) from text data (observed). In doing so, these applications choose a representation of text (text rep.) and a method to adjust for confounding.
    \label{tab:examples}}
\end{table*}

\vspace{-0.5cm}
\section{Applications}
\label{s:examples}
In Table~\ref{tab:examples}, we gather and summarize applications that use text to adjust for potential confounding. 
This encompasses both (a) text as a surrogate for confounders, or (b) the language itself as confounders.\footnote{ We acknowledge that Table~\ref{tab:examples} is by no means exhaustive. To construct Table~\ref{tab:examples}, we started with three seed papers: \citet{roberts2018adjusting}, \citet{veitch2019using}, and \citet{wood2018challenges}. We then examined papers cited by these papers, papers that cited these papers, and papers published by the papers' authors. We repeated this approach with the additional papers we found that adjusted for confounding with text. We also examined papers matching the query ``causal" or ``causality" in the ACL Anthology.} 

As an example, consider \citet{kiciman2018using} where
the goal is to estimate the size of the causal effect of alcohol use (treatment) on academic success (outcome) for college students. Since randomly assigning college students to binge drink
is not feasible or ethical,
the study instead uses observational data from Twitter,
which also has the advantage of a large sample size
of over sixty-three thousand students. They use heuristics to identify the Twitter accounts of college-age students and extract alcohol mentions and indicators of college success (e.g.,~study habits, risky behaviors, and emotions) from their Twitter posts. They condition on an individual's previous posts (temporally previous to measurements of treatment and outcome) as confounding variables since they do not have demographic data. They represent text as word counts and use stratified propensity score matching to adjust for the confounding bias. The study finds the effects of alcohol use include decreased mentions of study habits and positive emotions and increased mentions of potentially risky behaviors. 

\textbf{Text as a surrogate for confounders.}
Traditionally, causal research that uses human subjects as the unit of analysis would infer demographics via surveys. 
However, with the proliferation of the web and social media, social research now includes large-scale observational data that would be challenging to obtain using surveys \cite{salganik2017bit}. This type of data typically lacks demographic information but may contain large amounts of text written by participants from which demographics can be extracted. In this space, some researchers are specific about the confounders they want to extract such as an individual's ideology \cite{sridhar2019estimating} or mood \cite{sridhar2018mood}. Other researchers condition on all the text they have available and assume that low-dimensional summaries capture all possible confounders. For example, researchers might assume that text encodes all possible confounders between alcohol use and college success \cite{kiciman2018using} or psychiatric medication and anxiety \cite{saha2019social}. We dissect and comment on this assumption in Section~\ref{sec:conclusion}. 

\emph{\textbf{Open problems:}}
NLP systems have been shown to be inaccurate for low-resource languages \cite{Duong2015LowRD}, and exhibit racial and gender disparity \cite{blodgett2017racial,Zhao2017MenAL}. Furthermore, the ethics of predicting psychological indicators, such as mental health status, from text are questionable \cite{chancellor2019human}.
It is unclear how to mitigate these disparities when trying to condition on demographics from text and how NLP errors will propagate to causal estimates. 

\textbf{Language as confounders.}
There is growing interest in measuring language itself (e.g.~the sentiment or topical content of text) as causal confounders. For example, \citet{roberts2018adjusting} examine how the perceived gender of an author affects the number of citations that an article receives. However, an article's topics (the confounders) are likely to influence the perceived gender of its author (reflecting an expectation that women write about certain topics) and the number of citations of that article (``hotter" topics will receive more citations).
Other domains that analyze language as a confounder include news \cite{johansson2016learning}, social media \cite{de2016discovering,olteanu2017distilling}, and loan descriptions \cite{pham2017deep}. See Section~\ref{s:represent} for more discussion on the challenges and open problems of inferring these latent aspects of language. 

\vspace{-0.2cm}
\section{Estimating causal effects}
\label{s:causal-intro}
\vspace{-0.2cm}
Two predominant causal inference frameworks are  \emph{structural causal models (SCM)} \cite{pearl2009causality} and \emph{potential outcomes} \cite{rubin1974estimating,rubin2005causal},
which are complementary and theoretically connected  \cite{pearl2009causality,richardson2013single,morgan2015counterfactuals}. 
While their respective goals substantially overlap,
methods from structural causal models tend to emphasize
conceptualizing, expressing, and reasoning about the effects of possible causal relationships among variables, while methods from potential outcomes tend to emphasize
estimating the size or strength of causal effects. 

\vspace{-0.25cm}
\subsection{Potential outcomes framework} 
In the ideal causal experiment, for each each unit of analysis, $i$ (e.g.,~a person), one would like to measure the outcome, $y_i$ (e.g., an individual’s life expectancy), in both
a world in which the unit received treatment, $t_i =1$ (e.g.,~the person smoked), 
as well as in the counterfactual world in which the same unit did not receive treatment, $t_i=0$ (e.g~the same person did not smoke).\footnote{In this work, we only address binary treatments, but multi-value treatments
are also possible (e.g.,~\citet{imbens2000role}).}
A fundamental challenge of causal inference is that one cannot simultaneously observe treatment and non-treatment for a single individual 
\cite{holland1986statistics}. 

The most common population-level estimand of interest is the \emph{average treatment effect (ATE)}.\footnote{Other estimands include the average treatment effect on the treated (ATT) and average treatment effect on the control (ATC) \cite{morgan2015counterfactuals}} In the absence of confounders, this is simply the difference in means between the treatment and control groups, $\tau= \mathbbm{E}( y_i | t_i =1) - \mathbbm{E}(y_i | t_i = 0)$, and the ``unadjusted" or ``naive" estimator is 
\vspace{-0.1cm}
    \begin{equation}
    \hat{\tau}_{\text{naive}} = \frac{1}{n_1} \sum_{i: t_i=1} y_i -  \frac{1}{n_0} \sum_{j:t_j=0} y_j         \label{e:naive}
    \end{equation}
\vspace{-0.15cm}

\noindent
where $n_1$ is the number of units that have received treatment and $n_0$ is the number of units that have not received treatment. However, this equation will be biased if there are confounders, $z_i$, that influence both treatment and outcome. 

\subsection{Structural causal models framework} \label{s:scm}
\emph{Structural causal models} (SCMs) use a graphical formalism that
depicts nodes as random variables and directed edges as the direct causal dependence between these variables. 
The typical estimand of choice for SCMs is the probability distribution of an outcome variable $Y$ given an intervention on a treatment variable $T$:
\vspace{-0.1cm}
\begin{equation}
    P(Y \mid do(T=t)) \label{eqn:do}
    \vspace{-0.2cm}
\end{equation}
\noindent 
in which the \emph{do}-notation represents intervening to set variable $T$ to the value $t$ and thereby removing all incoming arrows to the variable $T$.

\textbf{Identification.} In most cases, Equation~\ref{eqn:do} is \emph{not} equal to the ordinary conditional distribution $P(Y \mid T=t)$ since the latter is simply filtering to the sub-population and the former is changing the underlying data distribution via intervention. Thus, for observational studies that lack intervention, one needs an \emph{identification strategy} in order to represent $P(Y \mid  do(T=t))$ in terms of distributions of observed variables. One such identification strategy (assumed by the applications throughout this review) is the \emph{backdoor criterion} which applies to a set of variables, $\mathcal{S}$, if they (i) block every backdoor path between treatment and outcome, and (ii) no node in $\mathcal{S}$ is a descendant of treatment. Without positive identification, the causal effects cannot be estimated and measuring variables from text is a secondary concern.

\begin{figure}[t]
\centering
\includegraphics[width=0.5\columnwidth]{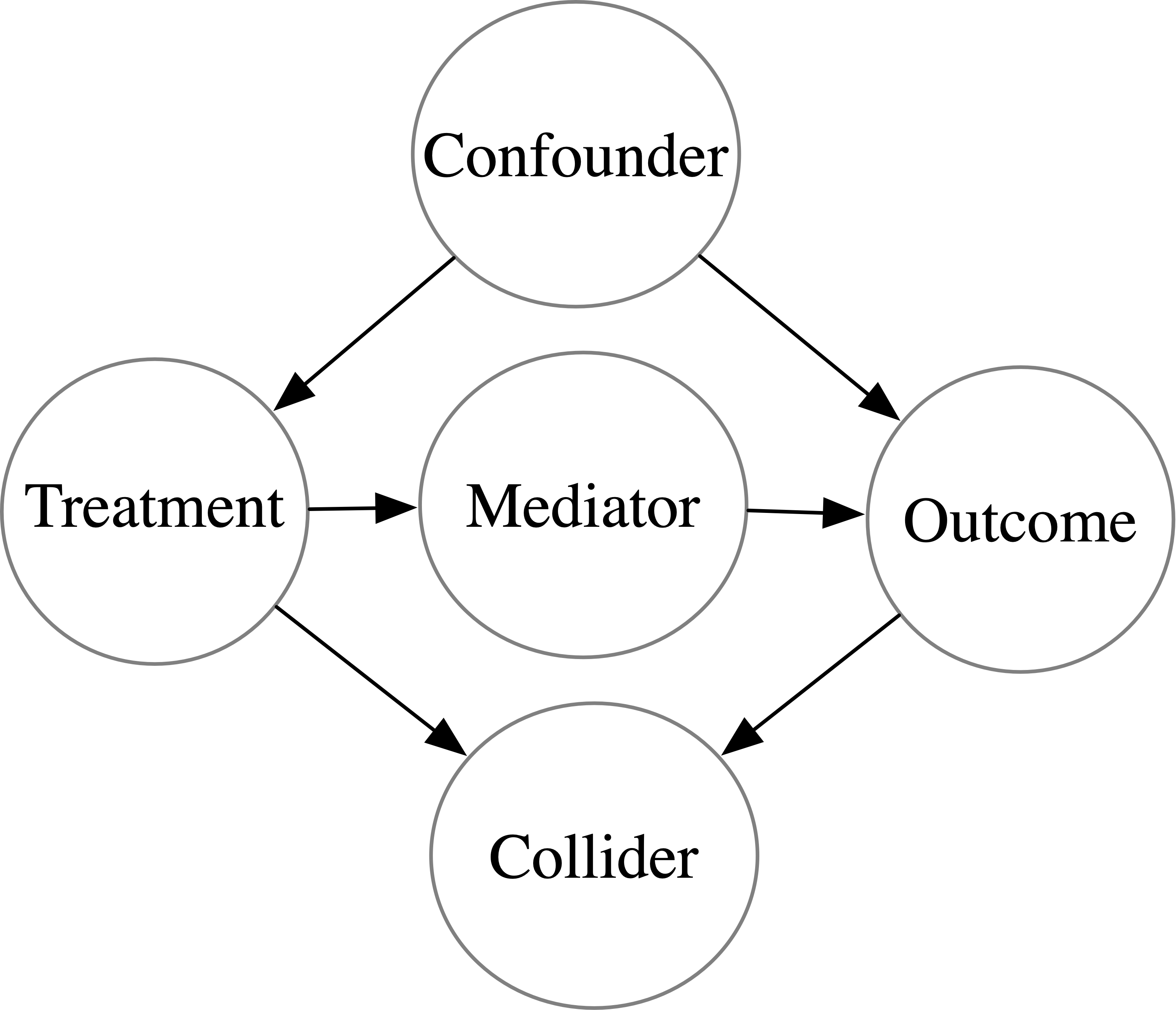}
\caption{A causal diagram showing common causal relationships.  
\label{f:causal-rel}}
\end{figure}

\textbf{Drawing the causal graph.}
Causal graphs help clarify which variables should and should not be conditioned on. 
The causal graphs in Figure~\ref{f:causal-rel} illustrate how the direction of the arrows differentiates confounder,
collider, and mediator variables.
Identifying the differences in these variables is crucial since, by \emph{d-separation}, conditioning on a confounder will block the  treatment-confounder-outcome path, removing bias.
By contrast, conditioning on a collider can create dependence between treatment-collider-outcome\footnote{
In \citet{pearl2016causal}'s example of a collider, suppose scholarships at a college are only given to two types of students: those with unusual musical talents and high grade point averages. In the general population, musical and academic talent are independent. However, if one discovers a person is on a scholarship (conditioning on the collider) then knowing a person lacks musical talent tells us that they are extremely likely to have a high GPA. } \cite{pearl2009causal} potentially introducing more bias \cite{montgomery2018conditioning,elwert2014endogenous}.
Mediator variables require a different set of adjustments than confounders to find the ``natural direct effect" between treatment and outcome \cite{vanderweele2015explanation,pearl2014interpretation}.
A practitioner typically draws a causal graph by explicitly encoding theoretical and domain assumptions as well as the results of prior data analyses.\footnote{See \citet{morgan2015counterfactuals} pgs.~33-34 on both the necessity and difficulty of specifying a causal graph for applied social research. \emph{Time-ordering} can be particularly helpful when encoding causal relationships (for instance, there cannot be an arrow pointing from variable $A$ to variable $B$ if $B$ preceded $A$ in time).}

\emph{\textbf{Open Problems:}} When could text potentially encode confounders and colliders simultaneously? If so, is it possible to use text to adjust exclusively for confounders?

\section{Measuring confounders via text} \label{s:represent}
After drawing the causal graph, the next step is to use available text data to recover latent confounders. Some approaches \emph{pre-specify} the confounders of interest and measure them from text, $P(z \mid x)$.  Others learn confounders \emph{inductively} and use a low-dimensional representation of text as the confounding variable $z$ in subsequent causal adjustments.  

\textbf{Pre-specified confounders.}
When a practitioner can specify confounders they want to measure from text (e.g.,~extracting ``occupation" from text in our smoking example), they can use either (1) \emph{lexicons} or (2) trained \emph{supervised classifiers} as the instrument of measurement.  Lexicons are word lists that can either be hand-crafted by researchers or taken  off-the-shelf. For example, \citet{saha2019social} use categories of the Linguistic Inquiry and Word Count (LIWC) lexicon \cite{pennebaker2001linguistic} such as tentativeness, inhibition, and negative affect, and use indicators of these categories in the text as confounders. Trained supervised classifiers use annotated training examples to predict confounders. For instance, \citet{saha2019social} also build machine learning classifiers for users' mental states (e.g.,~depression and anxiety) and apply these classifiers on Twitter posts that are temporally prior to treatment. If these classifiers \emph{accurately} recover mental states and there are no additional latent confounders, then conditioning on the measured mental states renders treatment independent of potential outcomes.

\emph{\textbf{Open problems:}}
Since NLP methods are still far from perfectly accurate, how can one mitigate error that arises from \emph{approximating} confounding variables? 
Closely related to this question is \emph{effect restoration} which addresses error from using proxy variables (e.g., a father's occupation) in place of true confounders (e.g, socioeconomic status)  \cite{kuroki2014measurement,oktay2019identifying}. \citet{wood2018challenges} build upon effect restoration for causal inference with text classifiers, but there are still open problems in accounting for error arising from other text representations and issues of calibration \cite{nguyen2015posterior} and prevalence estimation \cite{card2018importance,keith2018uncertainty} in conjunction with NLP. Ideas from the large literature on measurement error models may also be helpful \citep{fuller1987measurement,carroll2006measurement,buonaccorsi2010measurement}.

\textbf{Inductively derived confounders.}
Other researchers \emph{inductively} learn confounders in order to condition on \emph{all} aspects of text, known and unknown. For example, some applications condition on the entirety of news \cite{johansson2016learning} or scientific articles \cite{veitch2019using,roberts2018adjusting}.
This approach typically summarizes textual information with text representations common in NLP. Ideally, this would encode all aspects of language (meaning, topic, style, affect, etc.), though this is an extremely difficult, open NLP problem.
Typical approaches include the following.
(1) \emph{Bag-of-words} representations discard word order and use word counts as representations. (2) \emph{Topic models} are generative probabilistic models that learn latent topics in document collections and represent documents as distributions over topics \cite{blei2003latent,boyd2014care,roberts2014structural}. (3) \emph{Embeddings} are continuous, vector-based representations of text. To create vector representations of longer texts, off-the-shelf word embeddings such as \emph{word2vec} \cite{mikolov2013distributed} or \emph{GloVe} \cite{pennington2014glove} or combined via variants of weighted averaging \cite{arora2017simple} or neural models \cite{iyyer2015deep,bojanowski2017enriching,yang2016hierarchical}. (4) Recently, fine-tuned, large-scale neural language models such as BERT \cite{devlin-etal-2019-bert} have achieved state-of-the-art performance on semantic benchmarks, and are now used as text representations. Each of these text representations is a real-valued vector that is used in place of the confounder, $z$, in a causal adjustment method (\S\ref{s:estimators})

\emph{\textbf{Open problems:}} Estimates of causal effects are contingent on the ``garden of forking paths” of data analysis,  meaning any ``paths” an analyst did not take could have resulted in different conclusions \cite{gelman2013garden}. For settings with causal confounders from text, the first fork is the choice of representation (e.g.,~topic models or embeddings) and the second fork is the pre-processing and hyperparameter decisions for the chosen representations. 

We highlight that these decisions have been shown to alter results in predictive tasks. 
For instance, studies have shown that
pre-processing decisions dramatically change topic models \cite{denny2018text,schofield2017pulling};
embeddings are sensitive to hyperparameter tuning \cite{levy2015improving} and the construction of the training corpus \cite{antoniak2018evaluating}; and fine-tuned language model performance is
sensitive to random restarts \cite{phang2018sentence}.  Thus, reporting \emph{sensitivity analysis} of the causal effects from these decisions seems crucial:
how robust are the results to variations in modeling specifications?
 
\section{Adjusting for confounding bias} \label{s:estimators}
Given a set of variables $Z$ that satisfy the backdoor criterion (\S\ref{s:scm}), one can use the \emph{backdoor adjustment} to estimate the causal quantity of interest, 
\vspace{-.2cm}
\begin{align}
\begin{split}
    &P(Y=y \mid do (T=t) )\ = \\
    &\int\ P(Y=y \mid T=t,Z=z)\ P(Z=z)\  dz
\end{split}
\end{align}
\vspace{-.4cm}

\noindent
Conditioning on all confounders is often impractical in high-dimensional settings such as those found in natural language. We provide an overview of methods used by applications in this review that approximate such conditioning, leading to unbiased estimates of treatment effect; 
however, we acknowledge this is not an exhaustive list of methods and direct readers to more extensive guides \cite{morgan2015counterfactuals,athey2017estimating}. 

\emph{\textbf{Open problems:}} Causal studies typically make an assumption of  \emph{overlap}, also known as \emph{common support} or \emph{positivity}, meaning that any individual has a non-zero probability  of assignment to each treatment condition for all possible values of the covariates: $\forall z,\ \ 0<P(T=1 \mid Z=z)<1$. \citet{d2017overlap} show that as the dimensionality of covariates grows, strict overlap converges to zero. What are the implications of these results for high-dimensional text data?

\subsection{Propensity scores}
A \emph{propensity score} estimates the conditional probability of treatment given a set of possible confounders \cite{rosenbaum1984reducing,rosenbaum1983central,caliendo2008some}.
The true model of treatment assignment is typically unknown so one must estimate the propensity score from data (e.g.,~from a logistic regression model),  
\vspace{-.2cm}
\begin{equation}
{\pi} \equiv P(T =1 \mid Z).
\label{eqn:propensity-score}
\vspace{-.2cm}
\end{equation}

\noindent
\emph{Inverse Probability of Treatment Weighting (IPTW)} 
assigns a weight to each unit based on the propensity score \cite{lunceford2004stratification},
\begin{equation}
    w_i = 
    t_i/\hat{\pi}_i + 
    (1- t_i)/(1-\hat{\pi}_i), \label{e:weight}
    \vspace{-.2cm}
\end{equation}
thus emphasizing, for example, treated units that were originally unlikely to be treated ($t_i=1$, low $\pi_i$).
The ATE is calculated with weighted averages between
the treatment and control groups,\footnote{\citet{lunceford2004stratification} note there are two versions of IPTW, where both the weighted sum and the raw count have been used for the $n_0$ and $n_1$ denominators.} 
\vspace{-0.2cm}
\begin{equation}
    \hat{\tau}_{\text{IPTW}} = \frac{1}{n_1} \sum_{i : t_i =1} w_i y_i - \frac{1}{n_0} \sum_{j: t_j = 0} w_j y_j
\end{equation}
\vspace{-0.2cm}
\subsection{Matching and stratification}
\emph{Matching} aims to create treatment and control groups with similar confounder assignments;
for example, grouping units by observed variables (e.g.,\ age, gender, occupation),
then estimating effect size within each stratum \cite{stuart2010matching}.
\emph{Exact matching} on confounders is ideal but nearly impossible to obtain with high-dimensional confounders, including those from text. A framework for matching with text data is described by \citet{mozer2018matching} and requires choosing: (1) a text representation (\S\ref{s:represent}); (2) a distance metric (cosine, Eucliean, absolute difference in propensity score etc.); and (3) a matching algorithm. As \citet{stuart2010matching} describes, the matching algorithm involves additional decisions about (a) greedy vs.\ optimal matching; (b) number of control items per treatment item; (c) using calipers (thresholds of maximum distance); and (d) matching with or without replacement. \emph{Coarsened exact matching (CEM)} matches on discretized raw values of the observed confounders \cite{iacus2012causal}.

Instead of directly matching on observed variables, \emph{stratified propensity-score matching} partitions propensity scores into intervals (strata) and then all units are compared within a single strata \cite{caliendo2008some}. \emph{Stratification} is also known as interval matching, blocking, and subclassification. 

Once the matching algorithm is implemented, counterfactuals (estimated potential outcomes) are obtained from the matches $\mathcal{M}_i$ for each unit $i$:
\vspace{-0.2cm}
\begin{align}
    \hat{y}_i (k) = 
        \begin{cases} 
            y_i &\text{ if } t_i=k \\
        \frac{1}{|\mathcal{M}_i|}\sum_{j \in \mathcal{M}_i} y_j &\text{ if } t_i \not = k \\ 
     \end{cases} 
\end{align}
\vspace{-0.2cm}
\noindent which is plugged into the 
matching estimator,\footnote{For alternative matching estimators see \citet{abadie2004implementing}. This estimator is techinally the \emph{sample} average treatment effect (SATE), not the population-level ATE, since we have pruned treatment and control pairs that do not have matches \cite{morgan2015counterfactuals}.} 
\vspace{-.2cm}
\begin{equation}
    \hat{\tau}_{\text{match}} = \frac{1}{n} \sum_{i}^n \bigg( \hat{y}_i(1) - \hat{y}_i(0) \bigg). 
\end{equation}
\vspace{-.4cm}

\emph{\textbf{Open problems:}}
\citet{ho2007matching} describe matching as a method to reduce model dependence because, unlike regression, it does not rely on a parameteric form.  Yet, estimated causal effects may still be sensitive to other matching method decisions such as the number of bins in coarsened exact matching, the number of controls to match with each treatment in the matching algorithm, or the choice of caliper. Are causal estimates made using textual covariates particularly sensitive or robust to such choices?

\subsection{Regression adjustment}
\emph{Regression adjustment} fits a supervised model from observed data about the expected conditional outcomes
\vspace{-.3cm}
\begin{equation}
    q(t, z) \equiv \mathbbm{E}(Y \mid T=t, Z=z) \label{eqn:conditional-outcome} 
\end{equation}
\vspace{-.5cm}

\noindent
Then the learned conditional outcome, $\hat{q}$, is used to predict counterfactual outcomes for each observation under treatment and control regimes,
\vspace{-.2cm}
\begin{equation}
    \hat{\tau}_{\text{reg}} = \frac{1}{n} \sum_{i}^n \left(\hat{q}(1, z_i) - \hat{q}(0, z_i)\right) 
\end{equation}
\vspace{-0.8cm}
\subsection{Doubly-robust methods} 
Unlike methods that model only treatment (IPTW) or only outcome (regression adjustment), doubly robust methods model both treatment and outcome, and have the desirable property that if either the treatment or outcome models are unbiased then the effect estimate will be unbiased as well. These methods often perform very well in practice \citep{dorie2019automated}.
\emph{Adjusted inverse probability of treatment weighting (A-IPTW)} 
combines estimated propensity scores (Eqn.~\ref{eqn:propensity-score}) and 
conditional outcomes (Eqn.~\ref{eqn:conditional-outcome}),
while the more general
\emph{targeted maximum likelihood estimator (TMLE)} 
updates the conditional outcome estimate with a regression on the 
propensity weights (Eqn.~\ref{e:weight}) and $\hat{q}$ 
\cite{van2011targeted}. 

\subsection{Causal-driven representation learning}
Several research efforts design representations of text specifically for causal inference goals. These approaches still initialize their models with representations of text described in Section~\ref{s:represent}, but then the representations are updated with machine learning architectures that incorporate the observed treatment assignment and other causal information. 
\citet{johansson2016learning} design a network with a multi-task objective that aims for low prediction error for the conditional outcome estimates, $q$, and minimizes the discrepancy distance between $q(1, z_i)$ and $q(0, z_i)$ in order achieve balance in the confounders. 
\citet{roberts2018adjusting} combine structural topic models (STM; \citet{roberts2014structural}), propensity scores, and matching. They use the observed treatment assignment as the content covariate in the STM, append an estimated propensity score to the topic-proportion vector for each document, and then perform coarsened exact matching on that vector. \citet{veitch2019using} fine-tune a pre-trained BERT network with a multi-task loss objective that estimates (a) the original masked language-modeling objective of BERT, (b) propensity scores, and (c) conditional outcomes for both treatment and control. They use the predicted conditional outcomes and propensity scores in regression adjustment and the TMLE formulas. 

\emph{\textbf{Open problems:}} These methods have yet to be compared to one another on the same benchmark evaluation datasets. Also, when are the causal effects sensitive to hyperparameter and network architecture choices and what should researchers do in these settings?  

\vspace{-0.2cm}
\section{Human evaluation of intermediate steps}
\label{sec:eval-adjust}
Text data has the advantage of being \emph{interpretable}---matched pairs and some low-dimensional representations of text can be read by humans to evaluate their quality. When possible, we suggest practitioners use (1) interpretable balance metrics and/or (2) human judgements of treatment propensity to evaluate intermediate steps of the causal estimation pipeline. 

\vspace{-0.2cm}
\subsection{Interpretable balance metrics}
For matching and propensity score methods, the confounder balance should be assessed, since ideally 
$P(Z \mid T=1) = P(Z \mid T=0)$ in a matched sample \cite{stuart2010matching}. A standard numerical balance diagnostic is the \emph{standardized difference in means (SDM)},
\vspace{-0.2cm}
\begin{equation*}
    SDM(j) = \frac{\frac{1}{n_1}\sum_{i: t_i =1} z_{ij} - \frac{1}{n_0}\sum_{i: t_i =0} z_{ij}}{\sigma^{t=1}_j}
\end{equation*}
\noindent
where $z_{ij}$ is a single confounder $j$ for a single unit $i$ and  $\sigma^{t=1}_j$ is the standard deviation of $z_{ij}$ for all $i$ such that $t_i=1$.
SDM can also be used to evaluate the propensity score, in which case there would only be a single $j$ 
\cite{rubin2001using}. 

For causal text applications,  \citet{roberts2018adjusting} and \citet{sridhar2019estimating} estimate the difference in means for each topic in a topic-model representation of confounders and \citet{sridhar2018mood} estimate the difference in means across structured covariates but not the text itself. As an alternative to SDM, 
\citet{roberts2018adjusting} use string kernels to perform similarity checks.   
Others use domain-specific, known structured confounders to evaluate the balance between treatment and control groups. 
For instance, \citet{de2017language} sample treatment-control pairs across all propensity score strata and label the sampled text based on known confounders (in their case, from a previously-validated codebook of suicidal ideation risk markers).

\emph{\textbf{Open problems:}}
For embeddings and causally-driven representations, each dimension in the confounder vector $z$ is not necessarily meaningful. How can balance metrics be used in this setting? 

\subsection{Judgements of treatment propensity}
When possible, one can also improve validation by evaluating matched items (posts, sentences, documents etc.) to humans for evaluation. Humans can either (a) use a scale (e.g.,~a 1-5 Likert scale) to rate items individually on their propensity for treatment, or (b) assess similarity of paired items after matching. A simple first step is for analysts to do ``in-house" evaluation on a small sample (e.g.,~\citet{roberts2018adjusting}), but a larger-sample experiments on crowd-working platforms can also increase the validity of these methods (e.g.,~\citet{mozer2018matching}).


\textbf{\emph{Open problems:}} How can these human judgement experiments be improved and standardized?
Future work could draw from a rich history in NLP of evaluating representations of topic models and embeddings \cite{wallach2009evaluation,bojanowski2017enriching,schnabel-etal-2015-evaluation} and evaluating \emph{semantic similarity} \cite{cer2017semeval,bojanowski2017enriching,reimers2019sentence}.
\textbf{}
\section{Evaluation of causal methods} \label{s:eval-est}
Because the true causal effects in real-world causal inference are typically unknown, causal \emph{evaluation} is a difficult and open research question.
As algorithmic complexity grows, the expected performance of causal methods can be difficult to estimate theoretically \cite{jensen2019comment}. 
Other causal evaluations involve \emph{synthetic data}.  However, as \citet{gentzel2019case} discuss, synthetic data has no ``unknown unknowns" and 
many researcher degrees of freedom, which limits their effectiveness.
Thus, we encourage researchers to evaluate with \emph{constructed observational studies} or \emph{semi-synthetic datasets}, although measuring latent confounders from text increases the difficulty of creating realistic datasets that can be used for empirical evaluation of causal methods. 

\subsection{Constructed observational studies} Constructed observational studies collect data from both randomized and non-randomized experiments with similar participants and settings. Evaluations of this kind include job training programs in economics \cite{lalonde1986evaluating,glynn2013front}, 
advertisement marketing campaigns \cite{gordon2019comparison}, and education \cite{shadish2008can}. For instance, \citet{shadish2008can} randomly assign participants to a randomized treatment (math or vocabulary training) and non-randomized treatment (participants choose their own training). They compare causal effect estimates from the randomized study with observational estimates that condition on confounders from participant surveys (e.g.,~sex, age, marital status, like of mathematics, extroversion, etc.).

\emph{\textbf{Open problems:}} To extend \emph{constructed observational studies} to text data, one could build upon \citet{shadish2008can} and additionally (a) ask participants to write free-form essays of their past educational and childhood experiences and/or (b) obtain participants' public social media posts. Then causal estimates that condition on these textual representation of confounders could be compared to both those with surveys and the randomized settings. Alternatively, one could find observational studies with both real covariates and text and (1) randomize treatment conditional on the propensity score model (constructed from the covariates but not the text) and (2) estimate causal effect given only text (not the covariates). Then any estimated non-zero treatment effect is only bias. 

\subsection{Semi-synthetic datasets} Semi-synthetic datasets use real covariates and synthetically generate treatment and outcome, as in the 2016 Atlantic Causal Inference Competition \cite{dorie2019automated}. Several applications in this review use real metadata or latent aspects of text to simulate treatment and outcome: \citet{johansson2016learning} simulate treatment and outcome from two centroids in topic model space from newswire text; \citet{veitch2019using} use indicators of an article's ``buzzy" keywords;
\citet{roberts2018adjusting} use ``quantitative methodology" categories of  articles that were hand-coded by other researchers.

\emph{\textbf{Open problems:}} Semi-synthetic datasets that use real covariates of text seem to be a better evaluation strategy than purely synthetic datasets. However, with semi-synthetic datasets, researchers could be inadvertently biased to choose metadata that they know their method will recover. A promising future direction is a competition-style evaluation like \citet{dorie2019automated} in which one group of researchers generates a causal dataset with text as a confounder and other groups of researchers evaluate their causal methods without access to the data-generating process. 

\vspace{-0.2cm}
\section{Discussion and Conclusion}\label{sec:conclusion}
\vspace{-0.2cm}
Computational social science is an exciting, rapidly expanding discipline.
With greater availability of text data, alongside improved natural language processing models,
there is enormous opportunity to conduct new and more accurate
causal observational studies by controlling for
latent confounders in text. While text data ought to be as useful for measurement and inference as ``traditional" low-dimensional social-scientific variables,
combining NLP with causal inference methods requires tackling major open research questions. Unlike predictive applications, causal applications have no ground truth and so it is difficult distinguish modeling errors and forking paths from the true causal effects. 
 In particular, we caution against using all available text in causal adjustment methods \emph{without} any human validation or supervision, since one cannot diagnose any potential errors.
Solving these open problems, along with the others presented in this paper, would be a major advance for NLP as a social science methodology.

\section*{Acknowledgments}
The authors thank Sam Witty, Jacob Eisenstein, Brandon Stewart, Zach Wood-Doughty, Andrew Halterman, Laura Balzer, and members of the University of Massachusetts Amherst NLP reading group for helpful feedback, as well as the anonymous referees for detailed peer reviews.


\bibliography{bib}
\bibliographystyle{acl_natbib}

\end{document}